\documentclass[10pt,twocolumn,letterpaper]{article}

\usepackage[pagenumbers]{cvpr} 

\usepackage{graphicx}
\usepackage{amsmath}
\usepackage{amssymb}
\usepackage{booktabs}

\usepackage{multirow}
\usepackage{tabularx,verbatim}
\usepackage{xspace}
\usepackage{algorithm}
\usepackage{algorithmic}
\usepackage{setspace}
\usepackage{comment}
\usepackage{bbm}
\usepackage{enumitem}
\usepackage{colortbl}
\usepackage{color, xcolor} 
\usepackage[T1]{fontenc}
\usepackage{bbm}
\usepackage{dsfont}
\usepackage{bbding}

\newcommand{\yt}[1]{{\color{black} {#1}}}
\newcommand{\modify}[1]{{\color{black} {#1}}}

\definecolor{cvprblue}{rgb}{0.21,0.49,0.74}
\definecolor{arrow}{RGB}{184,140,240}
\definecolor{nose}{RGB}{211,161,66}

\usepackage[pagebackref,breaklinks,colorlinks,allcolors=cvprblue]{hyperref}

\def\paperID{5878} 
\def\confName{CVPR}
\def\confYear{2026}

\title{FG-Portrait: 3D Flow Guided Editable Portrait Animation}

\author{
Yating Xu$^1$ \qquad Yunqi Miao$^2$ \qquad Evangelos Ververas$^3$ \qquad Jiankang Deng$^3$ \qquad Jifei Song$^4$
\\
$^1$National University of Singapore \qquad
$^2$University of Warwick \\
$^3$Imperial College London \qquad
$^4$University of Surrey \\
{\tt\small \{xuyt98,ytswxb\}@gmail.com \qquad \{e.ververas16,j.deng16\}@imperial.ac.uk}
}

\begin{document}

\twocolumn[{
\maketitle
\begin{figure}[H]
\hsize=\textwidth
\centering
\includegraphics[scale=0.56]{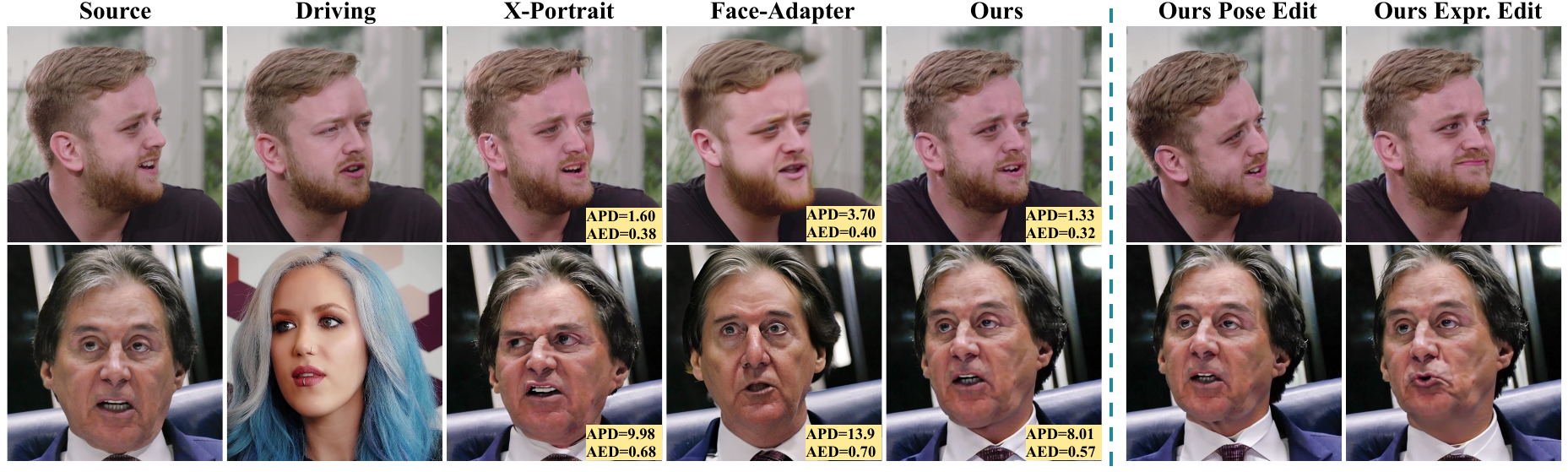}
\caption{Given a source and driving image, the goal of portrait animation is to synthesize the target image with source identity and driving head pose and expression. 
APD($\downarrow$) and AED($\downarrow$) denote the average pose and expression error compared to the driving pose and expression, respectively.
Compared to other diffusion-based models, \eg X-Portrait \cite{xie2024x} and Face-Adapter \cite{han2024face}, we can more accurately mimic the driving motion. Furthermore, we support feed-forward head pose and expression editing in the inference stage.
}
\label{teaser}
\end{figure}
}]

\begin{abstract}

\yt{Motion transfer from the driving to the source portrait remains a key challenge in the portrait animation.
Current diffusion-based approaches condition only on the driving motion,
which fails to capture source-to-driving correspondences and consequently yields suboptimal motion transfer. 
Although flow estimation provides an alternative, predicting dense correspondences from 2D input is ill-posed and often yields inaccurate animation.
We address this problem by introducing 3D flows, a learning-free and geometry-driven motion correspondence directly computed from parametric 3D head models.
To integrate this 3D prior into diffusion model, we introduce 3D flow encoding to query potential 3D flows for each target pixel to indicate its displacement back to the source location. 
To obtain 3D flows aligned with 2D motion changes,
we further propose depth-guided sampling to accurately locate the corresponding 3D points for each pixel.
Beyond high-fidelity portrait animation, our model further supports user-specified editing of facial expression and head pose.
Extensive experiments demonstrate the superiority of our method on consistent driving motion transfer as well as faithful source identity preservation. 
}
\end{abstract}    
\section{Introduction}
\label{sec:intro}

Portrait animation aims to reenact the person in a source image with the expression and head pose of a driving portrait. 
It has wide applications in film production, portrait editing and digital human reproduction.
However, faithfully transferring the driving motion poses a major challenge for portrait animation, especially under large pose or appearance variations between the source and driving portrait.


Inspired by the ability of diffusion models~\cite{rombach2022high} in generating high quality content,
recent works \cite{xie2024x,han2024face, ma2024follow, xu2025hunyuanportrait} condition on the driving motion for diffusion models to generate images with head pose and expression aligned with the driving portrait.
For example, Face-Adapter \cite{han2024face} extracts the landmark from reconstructed 3D head with the driving motion and source identity.
To improve the expressiveness of motion condition, X-Portrait~\cite{xie2024x} uses the driving image itself for motion control.
However, above motion conditions fail to provide connections between the source and driving portrait, leaving the learning of model ambiguous. 
Consequently, they lead to sub-optimal motion transfer (see the third and forth columns of Fig.~\ref{teaser}).

Existing attempts to establish motion correspondence between the source and driving images in portrait animation rely on a predicted motion field by the neural network \cite{hong2023implicit,doukas2023free,wang2021one,siarohin2019first, drobyshev2024emoportraits, drobyshev2022megaportraits}. These methods predict dense motion from 2D images, and then warp the source representation to follow the driving pose and expression. 
However, estimating the 3D movement from the 2D image is inherently ambiguous. 
Moreover, learning the motion field requires large scale training data, and can fail to generalize under large pose or appearance variations. 
Consequently, these approaches struggle to maintain consistent identity and realistic motion when the driving and source subjects differ significantly.

\begin{figure}[t]
\centering
\includegraphics[scale=0.57]{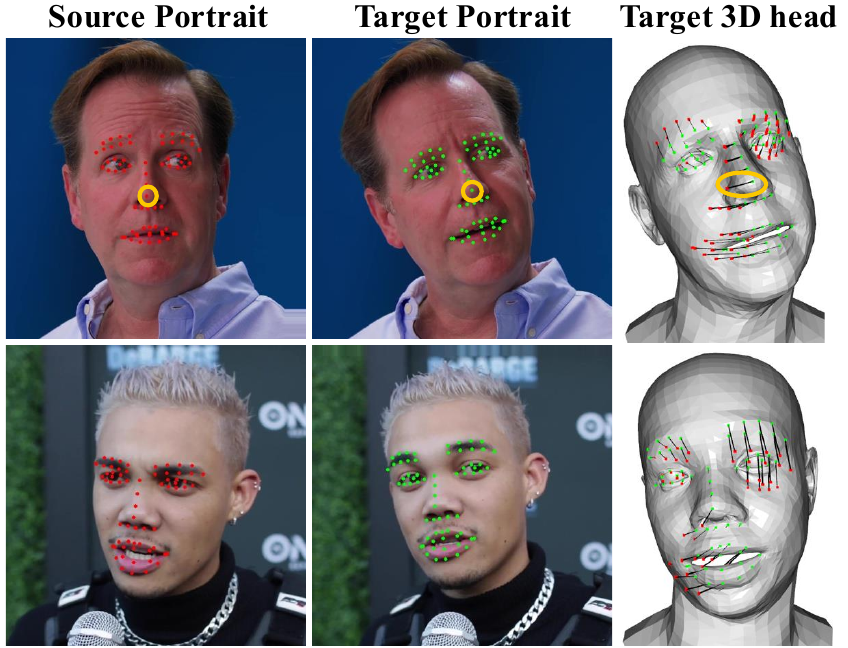}
\caption{
Visualization of 3D flows. Target portrait is the animated image with the source identity and driving motion.
Target 3D head is the head model of target portrait, which is pre-computed by assembling source shape and driving motion parameters.
We select some corresponding points both in the 2D and 3D, with \textcolor{red}{red} denoting the source and \textcolor{green}{green} denoting the target position. The 3D flows (black lines) correctly reflect the displacement from the target to the source position for each point. 
The yellow circles mark one example of a pair of points and the corresponding 3D flow.
}
\vspace{-5mm}
\label{3Dflow}
\end{figure}

In this paper, we introduce 3D flow as a learning-free and geometry-aware motion correspondence. It describes the 3D displacement between the source and driving motion on the parametric 3D head model.
Unlike previous methods that rely on network prediction, the proposed 3D flow is directly computed from the point-to-point semantic correspondence provided by the 3D head model.
As a result, the 3D flows correctly capture the movement between the source and driving portrait under divers motion change.
For example, the 3D flows in the first row of Fig.~\ref{3Dflow} faithfully demonstrate the per-point displacement when the person tilts his head to the right from the source to the target image. 
To incorporate the motion prior in portrait animation, 3D flow encoding is introduced as a new motion condition to diffusion model. We query the 3D flows along the backprojected ray for each pixel on the target image.
To align the 3D flow with corresponding movement in 2D images, we adopt a depth-guided sampling strategy where the backprojected 3D points are sampled based on the rendered depth map.

Our model further support feed-forward face editing for both expression and head pose based on the 3D parametric head model. 
During inference, the model can be driven not only by a driving image, but also by user-specified expression and pose parameters.
In summary, our contributions are as follows:
\begin{itemize}
\setlength{\parskip}{0.5mm} 
    \item \yt{We introduce 3D flow as a learning-free and geometry-aware motion correspondence for portrait animation. 
    It faithfully captures per-point driving-to-source 3D displacement across diverse pose and appearance variations.}
    \item We propose 3D flow encoding with depth-guided sampling as a new motion condition for the diffusion model, which queries 3D flows along the backprojected ray to establish accurate 3D$-$2D motion alignment.
    \item We further support user-specified expression and head pose editing in a feed-forward way during inference.
    \item We conduct extensive experiments on VFHQ and FFHQ datasets to demonstrate the superiority of our method qualitatively and quantitatively. 

\end{itemize}
\section{Related Work}

\begin{figure*}[t]
\centering
\includegraphics[scale=0.63]{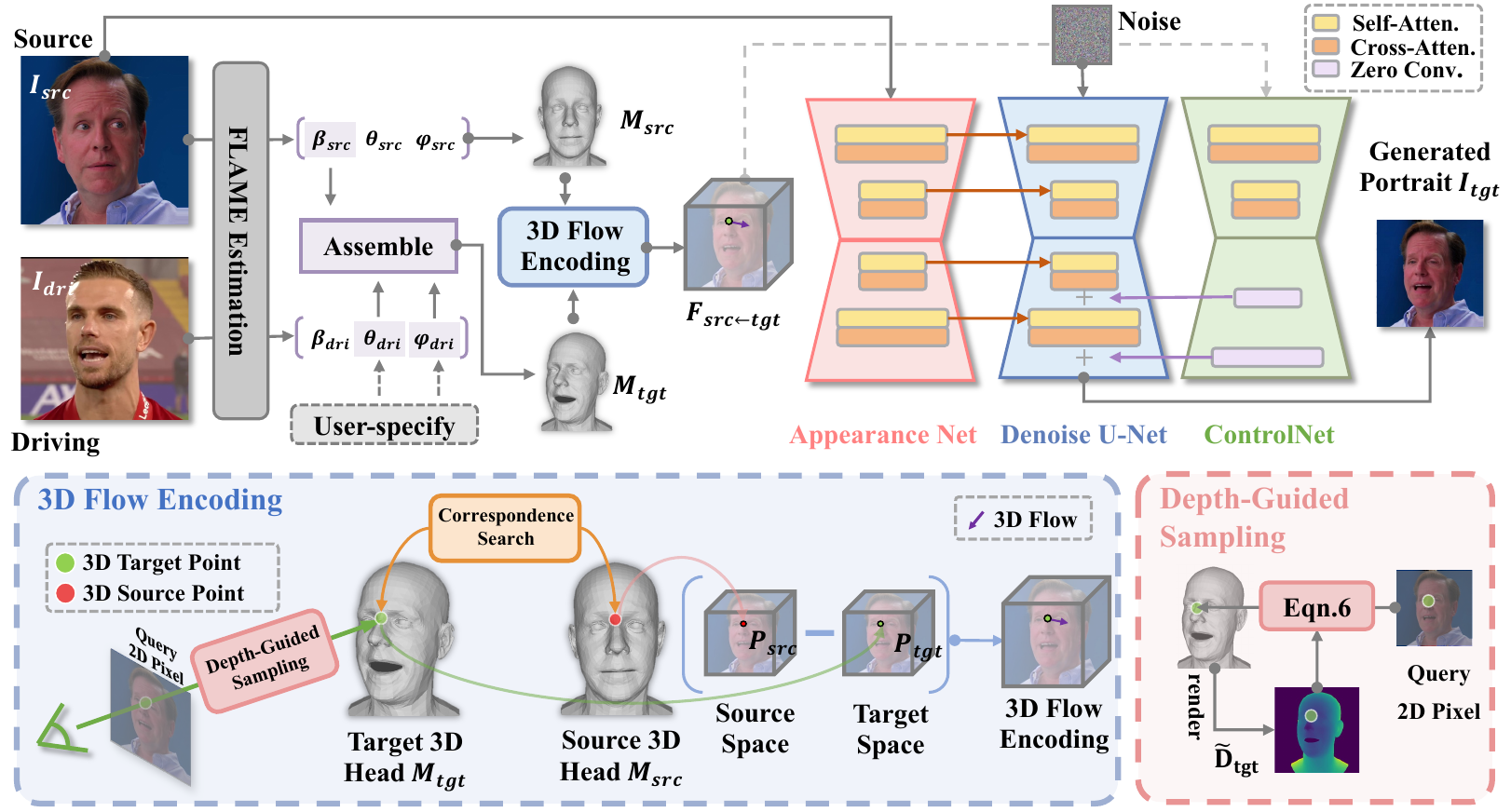}
\caption{Our Framework. 
We propose the 3D flow encoding $F_{src \leftarrow tgt}$ as the new motion condition to the ControlNet $G$.
Specifically, we first estimate the FLAME model $M_{tgt}$ and $M_{src}$ from $I_{dri}$ and $I_{src}$.
We perform depth-guided sampling to query the 3D flows in the target space for each pixel to indicate its displacement back to the source location, which are then stacked as the 3D flow encoding.
During inference, the driving motion $(\theta_{dri}, \psi_{dri})$ can come from the driving image or user-defined editing.
During training, $I_{dri}$ is sampled from the video of the  same person as $I_{src}$ and the training objective is to reconstruct $I_{tgt}$ as $I_{dri}$.
}
\label{method}
\end{figure*}

\label{sec:formatting}
\noindent \textbf{Diffusion Based Portrait Animation.}
Recently diffusion models have achieved superior performance in various generation tasks, such as image \cite{saharia2022photorealistic, nichol2021glide}, video \cite{guo2023animatediff, ho2022video, singer2022make} and 3D content \cite{poole2022dreamfusion, liu2023zero} generation.
Particularly, Stable Diffusion (SD) \cite{rombach2022high} is one of the most popular models, which shows unprecedented diversity and stability in the text-to-image generation.
Therefore, recent works \cite{xie2024x,ma2024follow,chang2023magicpose,hu2024animate,xu2024magicanimate, han2024face, xu2025hunyuanportrait, yang2025megactor} extend SD to portrait animation by adding detailed appearance and motion control.
For example, MagicPose \cite{chang2023magicpose} proposes an appearance network with multi-source attention to transfer the appearance of a human and the background from the source image to the generated image.  Motion control is achieved using ControlNet \cite{zhang2023adding}, with the driving human skeleton and facial landmarks as inputs.
Later works have mostly inherited the same appearance control mechanism as MagicPose, but made improvements in motion control, such as using expression-aware landmarks to avoid identity leakage \cite{ma2024follow} or replacing the coarse control signal of landmarks with the original driving image \cite{xie2024x}. 
Face-Adapter \cite{han2024face} extracts high-level identity features from the source image as the appearance condition and uses the landmarks of the source person performing the driving pose and expression as the motion condition.
Megactor-sigma \cite{yang2025megactor} and HunyuanPortrait \cite{xu2025hunyuanportrait} utilize more advanced generation backbone such as diffusion transformer \cite{peebles2023scalable} or stable video diffusion \cite{blattmann2023stable}, and adopt the latent representation of the driving image as the motion condition.
However, the motion condition in these methods fail to provide explicit motion links between the target and source instances.
As a result, the model has to independently establish this correspondence, often failing in challenging cases such as large pose variations or when the driving and source subjects differ.
In contrast, we propose the 3D flow encoding to densely describe the movement between the source and driving portraits, and therefore significantly ease the animation complexity.
Furthermore, superior to prior arts, we enable user-specified expression and head pose editing during inference.

\noindent \textbf{Motion Correspondence in Portrait Animation.}
There are some attempts to establish the motion correspondence between the source and the driving portrait in the portrait animation \cite{siarohin2019first, doukas2023free, zhao2022thin, wang2021one, hong2022depth, zeng2023face, guo2024liveportrait, drobyshev2024emoportraits, drobyshev2022megaportraits}. 
The pipeline consists of two stages, \ie, predicting motion flow map and warping the source features using the estimated flow map.
The training is end-to-end, with the flow map predicted in a self-supervised way.
The pioneering work FOMM \cite{siarohin2019first} first predicts sparse keypoints and local affine transformations for the source and driving images in a self-supervised way. Then, the dense motion flow is estimated in the feature space from the driving to the source frame based on the keypoints.
The source appearance features are warped by the flow map to generate the target image.
Follow-up works improve the flow estimation, for example, using thin-plate spline motion estimation to handle the non-linear complex motion  \cite{zhao2022thin} 
or replacing with 3D neural keypoints \cite{wang2021one} for head pose control. 
Other works predict the motion field from the latent motion representation and further enhance the rendering quality \cite{drobyshev2022megaportraits} and expression transfer \cite{drobyshev2024emoportraits} in a data-driven approach.
However, learning the motion field in a self-supervised way requires large dataset to train. Yet, it is error-prone when encountering large pose or appearance change.
In contrast, our method directly calculates the per-point motion flow using the aligned 3D head meshes.
The 3D parametric head model ensures that the estimated 3D flow faithfully reflects source-to-target motion, even under large pose and appearance variations.
\section{Method}

\noindent \textbf{Problem Definition.}
Given a source $I_{src} \in \mathbb{R}^{H \times W}$ and driving portrait $I_{dri} \in \mathbb{R}^{H \times W}$, the goal of portrait animation is to generate a target image $I_{tgt} \in \mathbb{R}^{H \times W}$, which retains the source identity and follows the same head pose and expression as $I_{dri}$.

\noindent \textbf{Overview.}
Fig.~\ref{method} shows our framework.
We follow existing works to implement separate appearance and motion controls for diffusion-based portrait animation (Sec.~\ref{preliminary}).
Particularly, we innovate the motion control by introducing \textbf{3D flow} (Sec.~\ref{subsec: 3d flow}) as motion correspondence, depicting the 3D displacement from the driving to the source motion state. 
To utilize 3D motion prior for animation, we further propose \textbf{3D flow encoding with depth-guided sampling} (Sec.~\ref{subsec: 3d flow encoding}) as new condition for the ControlNet, which encodes the corresponding 3D movement of each pixel in the target image moving back to the source location.
Finally, we show that we support user-specified editing during inference (Sec.~\ref{editing}).

\subsection{Preliminary}
\label{preliminary}
\noindent \textbf{Diffusion-based Portrait Animation.}
We follow existing works \cite{chang2023magicpose,xie2024x} to construct three branches for diffusion-based portrait animation.
The first branch is the image generator, which is instantiated  as the Stable Diffusion U-Net \cite{rombach2022high}, denoted as $U$. It iteratively denoises the Gaussian noise through $T$ time steps. During training, an image is first encoded in the latent space and further added with noise $\epsilon \sim \mathcal{N}(0,1)$ to \textit{t} time steps.
Then, $U$ learns to predict the added noise conditioned on the time step and other conditions as follows:
\begin{equation}
\label{loss}
\mathcal{L}=\mathbb{E}_{z_0, c, \epsilon , t}\left[\left\|\epsilon-U\left(z_t, t, c\right)\right\|_2^2\right],
\end{equation}
where $z_{t}$ is the noisy version of the image latent features and $c$ represents the extra conditions.
The second branch entails the appearance control. 
The appearance network $A$ has the same structure as $U$ and extracts the subject's appearance and background context from $I_{src}$. 
The extracted features $A(I_{src})$ serve as extra key-values pairs to modulate the self-attention layers in $U$.
The third branch is responsible for the motion control. 
Driving motion representation $R_{dri}$, such as landmarks \cite{chang2023magicpose} or original images \cite{xie2024x} are extracted from $I_{dri}$ and fed into a ControlNet $G$ \cite{zhang2023adding}.
The optimization goal is expressed by Eqn.~\ref{loss} with $c = (A(I_{src}), G(R_{dri}))$, where $I_{src}$ and $R_{dri}$ are the appearance and motion conditions, respectively.

\begin{figure*}[t]
\centering
\includegraphics[scale=0.53]{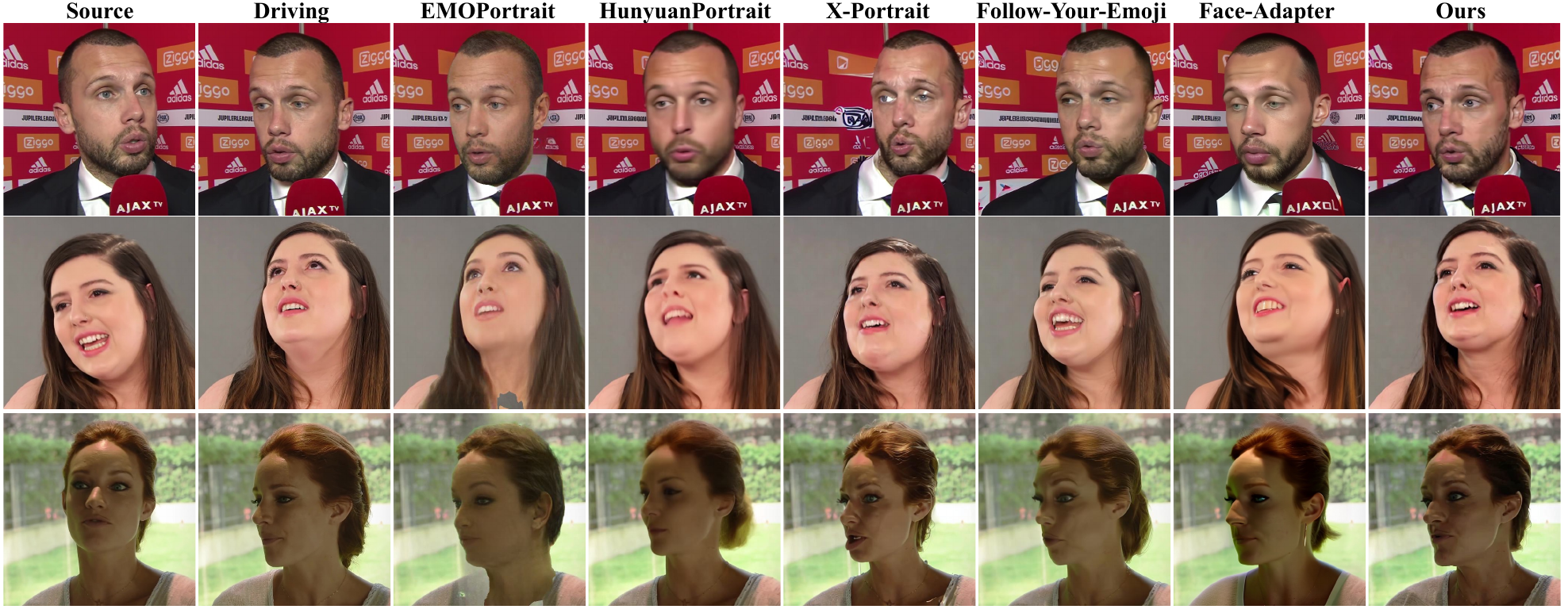}
\caption{Qualitative results on self-reenactment. ``Source'' and ``Driving'' denote the source and driving image, respectively.}
\label{self-figure}
\end{figure*}

Although diffusion-based methods can well maintain the source human appearance and background in the $I_{tgt}$, they tend to show unsatisfactory motion transfer especially under large motion or appearance variations (see Fig.~\ref{self-figure} and Fig.~\ref{cross-figure}).
Therefore, we introduce 3D flow encoding as the \textit{new} motion condition signal, which depicts the 3D displacement from the driving to the source motion state.

\begin{figure}[t]
\centering
\includegraphics[scale=0.65]{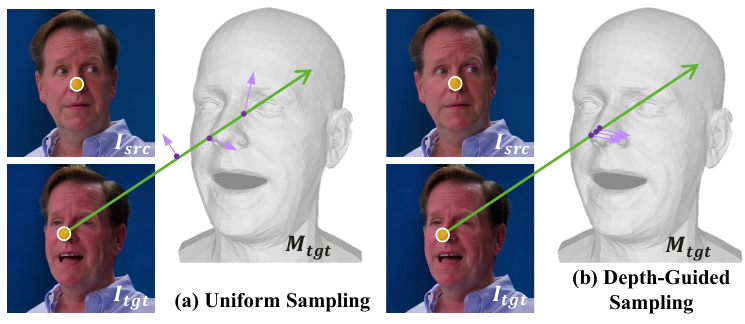}
\caption{Illustration of depth-guided sampling.
\textcolor{arrow}{Arrows} show the 3D flows for the selected pixel marked with \textcolor{nose}{yellow circle}. 
The sampled points in (b) are closer to the actual 3D location of the pixel than (a), resulting in 3D flows that are more faithfully aligned with the 2D motion.
}
\label{depth-guided}
\vspace{-3mm}
\end{figure}

\noindent \textbf{FLAME Model.}
The FLAME model \cite{li2017learning} is a parametric 3D head model with shape, pose and expression components. It can be described through a mapping function $M(\beta, \theta, \psi): \mathbb{R}^{|\beta| \times|\theta| \times|\psi|} \rightarrow \mathbb{R}^{L \times 3}$ with $L=5023$ vertices.
$\beta = [\beta^1, ..., \beta^{300}]^{\top} \in \mathbb{R}^{300}$ 
is the shape coefficients, which controls the identity-related shape variations through the shape blendshapes $\text{B}_\text{S}(\beta,S )$. $S = [s^1, ..., s^{300}]^{\top}$ denotes the shape basis, where each shape basis $s^n$ represents a unique head shape and $\beta^{n}$ controls the magnitude of $s^n$.
Similarly, $\psi = [\psi^1, ..., \psi^{100}]^{\top} \in \mathbb{R}^{100}$ is the expression coefficients, which controls the variations in expression through linear blendshapes $\text{B}_\text{E}(\psi,E)$. $E = [e^1, ..., e^{100}]^{\top}$ denotes the expression basis, where $e^n$ represents a unique expression type, \eg, anger or smile, and $\psi^n$ controls the magnitude of the expression $e^n$.
The FLAME pose vector $\theta = [\theta^1, ..., \theta^{12}]^{\top} \in \mathbb{R}^{12}$ describes the rotation of the neck, jaw and eyes in the axis-angle representation.
$M$ is rotated according to $\theta$ though linear blend skinning and further adjusted by the pose correctives.
Changing $\beta$, $\theta$ and $\psi$ generates diverse 3D human heads.

\subsection{3D Flow}
\label{subsec: 3d flow}
We introduce 3D flows to describe per-point 3D displacement between the driving and the source motion based on the parametric FLAME model.
Specifically, we first estimate the source FLAME $M_{src}$ and driving FLAME $M_{dri}$ from the $I_{src}$ and $I_{dri}$ as follow:
\begin{equation}
\begin{aligned}
M_{src} &= M(\beta_{src}, \theta_{src}, \psi_{src}) \\
M_{dri} &= M(\beta_{dri}, \theta_{dri}, \psi_{dri}),
\end{aligned}
\end{equation}
Then, the target FLAME $M_{tgt}$ is obtained by assembling the source shape coefficient $\beta_{src}$ with driving pose $\theta_{dri}$ and expression coefficient $\psi_{dri}$ as follows:
\begin{equation}
    M_{tgt} = M(\beta_{src}, \theta_{dri}, \psi_{dri}).
\end{equation}
Point-wise correspondence can be established semantically between $M_{src}$ and $M_{tgt}$ since each vertex of the FLAME corresponds to certain facial structure.
Specifically, given a point $p_{tgt} \in \mathbb{R}^3$ in the target space, we search its corresponding location $p_{src} \in \mathbb{R}^3$ in the source space using surface field SF \cite{bergman2022generative} as follow:
\begin{equation}
\label{sf}
    p_{src} = \operatorname{SF}\left(p_{tgt}; M_{tgt}, M_{src}\right).
\end{equation}
The SF assigns each $p_{tgt}$ to its nearest triangle face on $M_{tgt}$, and then computes the $p_{src}$ based on the matched source triangle face on the $M_{src}$.
Here, we conduct correspondence search in a backward fashion, \ie find $p_{src}$ given $p_{tgt}$, so that each target point is guaranteed to find its correspondence in the source space.
Then, the 3D flow $f_{{src} \leftarrow tgt} \in \mathbb{R}^3$ for $p_{tgt}$ depicting its displacement to $p_{src}$ can be computed as follows:
\begin{equation}
\label{flow}
    f_{src \leftarrow tgt} = p_{src} - p_{tgt}.
\end{equation}

The visualization of 3D flow is shown in the last column of Fig.~\ref{3Dflow}. The \textcolor{green}{green} and \textcolor{red}{red} dots represent \textcolor{green}{$p_{tgt}$} and \textcolor{red}{$p_{src}$}, respectively. The flows are visualized by the black lines connecting $p_{tgt}$ and corresponding $p_{src}$, which reflect the movement from the target back to the source head pose and expression.

\begin{table*}[t]
\begin{minipage}{0.5\linewidth}
\centering
\caption{Comparison on the self-reenactment task on VFHQ at $512^2$. $\downarrow$ means lower the better and $\uparrow$ is the opposite. APD is scaled by 100. Best results are marked bold.}
\label{sel-reenactment}
\resizebox{\linewidth}{!}{
\begin{tabular}{ccccc}
\toprule
Method       & LPIPS $\downarrow$ & CSIM $\uparrow$    & APD $\downarrow$ & AED $\downarrow$ \\ \midrule
EMOPortrait \cite{drobyshev2024emoportraits} &  0.235 & 0.729   & 3.047   & 0.371   \\
X-Portrait \cite{xie2024x}   & 0.195  & 0.777  & 3.660     & 0.357    \\ 
\small{Follow-Your-Emoji \cite{ma2024follow}} & 0.162 & 0.774 & 3.570  & 0.402 \\
Face-Adapter \cite{han2024face} & 0.222 & 0.699 & 3.386 & 0.400 \\
HunyuanPortrait \cite{xu2025hunyuanportrait} & \modify{0.162} & \modify{0.781} & \modify{3.440} & 0.341 \\
Ours & \textbf{0.158} & \textbf{0.807} & \textbf{2.682} & \textbf{0.327} \\
\bottomrule
\end{tabular}
}
\end{minipage}
\hspace{0.2cm}
\begin{minipage}{0.48\linewidth}
\centering
\caption{Comparison on the cross-reenactment task on VFHQ at $512^2$. $\downarrow$ means lower the better and $\uparrow$ is the opposite. APD is scaled by 100. Best results are marked bold.}
\label{cross-reenactment}
\resizebox{\linewidth}{!}{
\begin{tabular}{ccccc}
\toprule
Method       & FID $\downarrow$ & CSIM $\uparrow$   & APD $\downarrow$ & AED $\downarrow$ \\ \midrule
EMOPortrait \cite{drobyshev2024emoportraits} & 100.6 & 0.386   & \modify{7.860}   & \modify{0.660}   \\
X-Portrait \cite{xie2024x}   & 104.6  & 0.477  & 11.220     & 0.766    \\ 
\small{Follow-Your-Emoji \cite{ma2024follow}} & 91.3 & \textbf{0.484} & 9.291  & 0.788 \\
Face-Adapter \cite{han2024face} & 94.6 & 0.424 & 7.785 & 0.688 \\
HunyuanPortrait \cite{xu2025hunyuanportrait} & \modify{92.7} & 0.455 & 9.220 & \modify{0.658} \\
Ours & \textbf{87.0} & 0.462 & \textbf{7.764} & \textbf{0.652} \\
\bottomrule
\end{tabular}
}
\end{minipage}
\end{table*}

\subsection{3D Flow Encoding}
\label{subsec: 3d flow encoding}
To utilize the 3D flow for the 2D animation task, we propose 3D flow encoding $F_{src \leftarrow tgt} \in \mathbb{R}^{H\times W\times 3N}$.
We first query the potential 3D flows for each pixel, and then stack them as the motion condition for the diffusion model. 

For each pixel in target image, we backproject it to the target space and sample $N$ points along the ray as its possible 3D locations.
The $n$-th point $p_{tgt}^n$ is computed by sampling at depth $d_n$ in the target space as follow:
\begin{equation}
\label{backprojection}
p_{tgt}^n = H \left[d_n\left(K^{-1} q_{tgt}\right)^{\top}, 1\right]^{\top},
\end{equation}
where $q_{tgt}= (u, v, 1)^{\top}\in \mathbb{R}^3$ is the homogeneous coordinate of the pixel.
$K \in \mathbb{R}^{3 \times 3}$ and $H \in \mathbb{R}^{3 \times 4}$ are the camera intrinsic and extrinsic, respectively. 
Then, we stack the $N$ points at each pixel location, forming the 3D target positions $P_{tgt} \in \mathbb{R}^{H\times W\times N \times 3}$ for the target image.
Then, we search the corresponding source locations $P_{src} \in \mathbb{R}^{H\times W\times N \times 3}$ using Eqn.~\ref{sf}.
The 3D flow encoding $F_{src \leftarrow tgt} $ is computed by $F_{src\leftarrow tgt} = P_{src} - P_{tgt}$, which 
depicts the possible 3D displacements for each pixel between $I_{src}$ and $I_{tgt}$.
We reshape $F_{src \leftarrow tgt}$  into $(H\times W \times 3N)$ and input it into the ControlNet for motion control.

\noindent \textbf{Depth-Guided Sampling}
The location where the 3D flow is queried is important as the the 3D flow encoding should reflect the corresponding 2D motion change.
A straightforward way is to uniformly sampling flows at a pre-defined depth range along the backprojected ray as shown in the (a) of Fig.~\ref{depth-guided}. However, the 3D flows at these points hardly reflect the 2D movement, and consequently leads to inaccurate motion transfer (see Tab.~\ref{ablate-depth}). 
To this end, we propose depth-guided sampling to sample the flows near the corresponding 3D point for each pixel. 
Specifically, we estimate the depth map $\tilde{D}_{tgt} \in \mathbb{R}^{H \times W}$ of the target head as follow:
\begin{equation}
    \tilde{D}_{tgt} = \operatorname{Render}\left(M_{tgt}; H,K \right),
\end{equation}
where $\tilde{D}_{tgt}$ only estimates the depth of the head region while the areas outside the head are assigned depth value of 0.
Then, $d_n$ at Eqn.~\ref{backprojection} is sampled from the range \([\tilde{D}_{tgt}[u,v] - \delta, \tilde{D}_{tgt}[u,v] + \delta]\) if $q_{tgt}$ is within the projected head region. 
Otherwise, $d_n$ is sampled from a pre-defined depth range $[d_{near}, d_{far}]$ since these areas may still contain the areas related to the human, \eg, hair or hat.
Consequently, the 3D flow encoding correctly reflects the 2D movement as shown in the (b) of Fig.~\ref{depth-guided}, which provides faithful motion guidance during animation.

\noindent{\textbf{Objectives.}}
Our final optimization objective is Eqn.~\ref{loss} with motion condition updated as $F_{src \leftarrow tgt}$.





\subsection{Expression and Head Pose Editing}
\label{editing}
During inference, we further support user-specified motion editing on expression and head pose with varied intensity.
Let $\Delta\psi_{usr} \in \mathbb{R}^{100} $ and $\Delta\theta_{usr}  \in \mathbb{R}^{12}$ be the user-specified FLAME expression and pose editing parameters, respectively.
The final expression and pose coefficients in the $M_{tgt}$ is given as follows:
\begin{equation}
\begin{aligned}
    \psi_{\text{dri}} &\leftarrow \psi_{\text{dri}} + \Delta\psi_{\text{usr}} ,\\
    \theta_{\text{dri}} &\leftarrow \theta_{\text{dri}} + \Delta\theta_{usr}.
\end{aligned}
\end{equation}
Then, the 3D flow encoding $F_{src \leftarrow tgt}$ is updated accordingly based on the new $M_{tgt}$ and further sent to the $G$ to generate the new edited image.
\section{Experiments}

\subsection{Dataset and Setup}
\paragraph{Dataset.}
We conduct experiments mainly on the VFHQ \cite{xie2022vfhq} dataset, which contains high-fidelity human interview clips. We sample 1K videos as the training dataset. 
We use the official test split of VFHQ as the testing dataset.
We design two experimental settings, \ie, self-reenactment and cross-reenactment, as the testing benchmarks.
For the self-reenactment task, the source and the driving images are sampled from the same video. Specifically, we use one frame in every video as the source image, and uniformly sample another 10 frames from the same video as driving images.
For the cross-reenactment task, the source and driving images are from videos capturing different identities. We use one frame of a video as the source image while sampling 10 frames from the video of the other identity as the driving frames. In addition, we test the model's generalization on the FFHQ dataset \cite{karras2019style}. We randomly sample 100 images from the FFHQ dataset as source frames and use the VFHQ videos as the driving sequence. 

\noindent \textbf{Metrics.} 
In the self-reenactment experiment, we use LPIPS to evaluate the image quality, and CSIM \cite{deng2019arcface} between the $I_{tgt}$ and $I_{dri}$ to evaluate the identity preservation.
In the cross-reenactment experiment, given the absence of the ground-truth target image,
we use FID to evaluate the image quality and CSIM between the $I_{tgt}$ and $I_{src}$ to evaluate the identity preservation.
Both self and cross reenactment adopt the average pose distance (APD) \cite{retsinas20243d} and average expression distance (AED) \cite{retsinas20243d} to evaluate the pose and expression accuracy with respect to the $I_{dri}$, respectively.

\begin{table}[t]
\centering
\caption{Comparison on the cross-reenactment task on FFHQ at $512^2$. $\downarrow$ means lower the better and $\uparrow$ is the opposite. APD is scaled by 100. Best results are marked bold.}
\vspace{-2mm}
\label{ffhq}
\resizebox{\linewidth}{!}{
\begin{tabular}{ccccc}
\toprule
Method       & FID $\downarrow$ & CSIM $\uparrow$   & APD $\downarrow$ & AED $\downarrow$ \\ \midrule
EMOPortrait \cite{drobyshev2024emoportraits} & 105.1 & 0.384   & \modify{9.840}   & \modify{0.719}   \\
X-Portrait \cite{xie2024x}   & 131.4  & \textbf{0.654}  & 14.696     & 0.791    \\ 
\small{Follow-Your-Emoji \cite{ma2024follow}} & 120.3 & 0.634 & 10.624  & 0.863 \\
Face-Adapter \cite{han2024face} & 135.1 & 0.437 & 10.869 & 0.744 \\
HunyuanPortrait \cite{xu2025hunyuanportrait} & 113.7 & 0.505 & \modify{10.570} & \modify{0.722} \\
Ours & \textbf{99.4} & 0.558 & \textbf{9.297} & \textbf{0.714} \\
\bottomrule
\end{tabular}
}
\vspace{-3mm}
\end{table}

\begin{table}[t]
\centering
\caption{Comparison of motion conditions. 
`S-APD' and `S-AED' are the APD and AED for self-reenactment task. `C-APD' and `C-AED' are the APD and AED for cross-reenactment task.}
\vspace{-2mm}
\label{ablate-motion}
\resizebox{\linewidth}{!}{
\begin{tabular}{ccccc}
\toprule
Model     & S-APD $\downarrow$ & S-AED $\downarrow$    & C-APD $\downarrow$ & C-AED $\downarrow$ \\ \midrule
Dri-Ldk & 4.001 & 0.373	& 8.588 & 0.688 \\
Predicted Flow & 4.232 & 0.384	& 12.430 & 0.778 \\
Ours & \textbf{2.682} & \textbf{0.327} & \textbf{7.764} &	\textbf{0.652} \\
\bottomrule
\end{tabular}
}
\vspace{-3mm}
\end{table}

\begin{figure*}[t]
\centering
\includegraphics[scale=0.53]{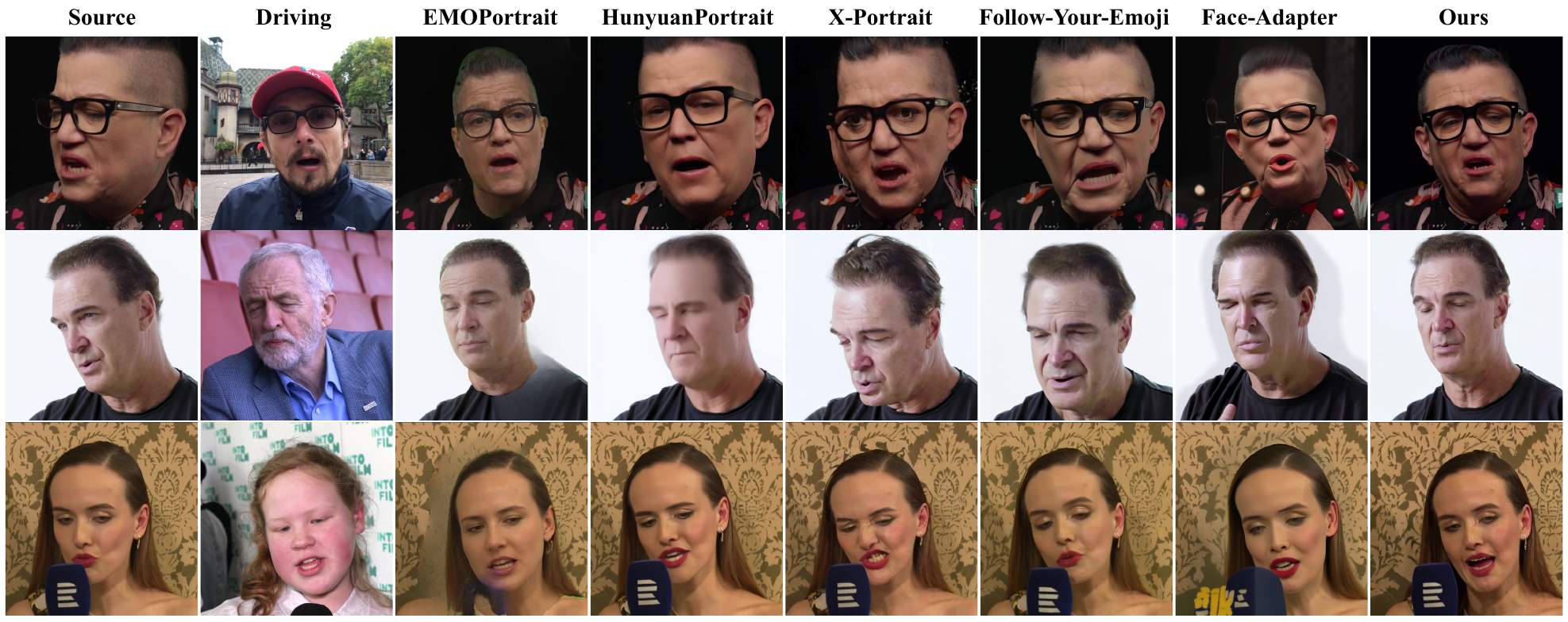}
\caption{Qualitative results on cross-reenactment. ``Source'' and ``Driving'' denote the source and driving image, respectively.}
\label{cross-figure}
\end{figure*}

\subsection{Implementation details}
The FLAME coefficients ($\beta$, $\theta$ and $\psi$) and camera parameters ($K$ and $H$) are obtained by running FLAME fitting method on each image following \cite{deng2024portrait4d}.
We sample $N=20$ points to construct the 3D flow encoding $F_{src \leftarrow tgt}$.
$\delta$ is set to 0.01m. $d_{far}$ and $d_{near}$ is set as $\pm 0.65$m to the world origin.
We use SD 1.5 as our generative backbone and we freeze its weights during training.
For the appearance net, we initialize its weights from X-portrait. For motion ControlNet, we prepend an additional input layer to match the input feature dimension with the original dimension in the ControlNet. The new layer is initialized randomly while the rest layers are initialized from weights of \cite{zhang2023adding}.
We jointly fine-tune the appearance net and the motion ControlNet using AdamW optimizer with learning rate  $1e^{-5}$.
Following previous work \cite{xie2024x, ma2024follow}, we further insert temporal layers into the diffusion model after finishing training image diffusion pipeline, and then finetune the temporal layers on the video sequence to achieve temporal consistency.

\subsection{Comparison with Baselines}
We compare with EMOPortrait \cite{drobyshev2024emoportraits}, X-Portrait \cite{xie2024x}, Follow-Your-Emoji \cite{ma2024follow}, Face-Adapter \cite{han2024face} and HunyuanPortrait \cite{xu2025hunyuanportrait}.
EMOPortrait is the state-of-the-art method which predicts motion field between the source and driving portrait. 
X-Portrait, Follow-Your-Emoji, Face-Adapter and HunyuanPortrait are the state-of-the-art diffusion-based portrait animation methods, which leverage driving image, facial landmarks or latent representation as the motion condition, respectively. 
All models generate images of size $512 \times 512$.

Tab.~\ref{sel-reenactment} and Tab.~\ref{cross-reenactment} show the results of self- and cross-reenactment on the VFHQ test dataset, respectively.
Due to the lack of source-to-driving correspondence in the motion condition, previous diffusion-based methods (X-portrait, Follow-Your-Emoji, Face-Adapter and HunyuanPortrait) show higher APD and AED error.
\yt{EMOPortrait \cite{drobyshev2024emoportraits} also shows inferior animation performance due to their inaccurate motion flow prediction and inferior generative backbone (GAN).}
Regarding identity preservation (CSIM), we note that our method is slightly lower than the best-performing baseline in Tab.~\ref{cross-reenactment}. This is expected because portrait animation inherently involves a trade-off between identity preservation and motion accuracy in the cross-reenactment. 
A model can easily achieve a high CSIM score by simply copying the source image, but this trivial solution leads to extremely large APD and AED as the driving motion is ignored.
Conversely, replicating the driving image can minimize APD and AED, but drastically reduces CSIM due to the loss of identity information.
Our approach strikes a better balance, yielding significantly lower APD and AED while maintaining a competitively high CSIM, demonstrating that our 3D flow enables accurate motion transfer without compromising source identity.

We further provide qualitative results in Fig.~\ref{self-figure} and Fig.~\ref{cross-figure}.
Previous diffusion-based methods show inferior motion transfer results under large pose variations or when the source and driving identities are different.
Under large motion in the driving sequence, HunyuanPortrait tends to produce over-smoothed results, as its powerful video backbone favors temporal consistency over motion fidelity.
\yt{Although EMOPortrait can roughly follow the driving pose and expression, its results suffer from poor identity preservation and inconsistent foreground–background blending. This stems from its design that animates only the segmented head region while keeping the background fixed. While this simplification eases motion generation, it limits visual fidelity and overall realism.
}
In contrast, our method models both the subject and background jointly within the diffusion framework, allowing the background to adapt coherently to the driving motion and preserving spatial consistency across the entire image.

\begin{figure*}[t]
\centering
\includegraphics[scale=0.53]{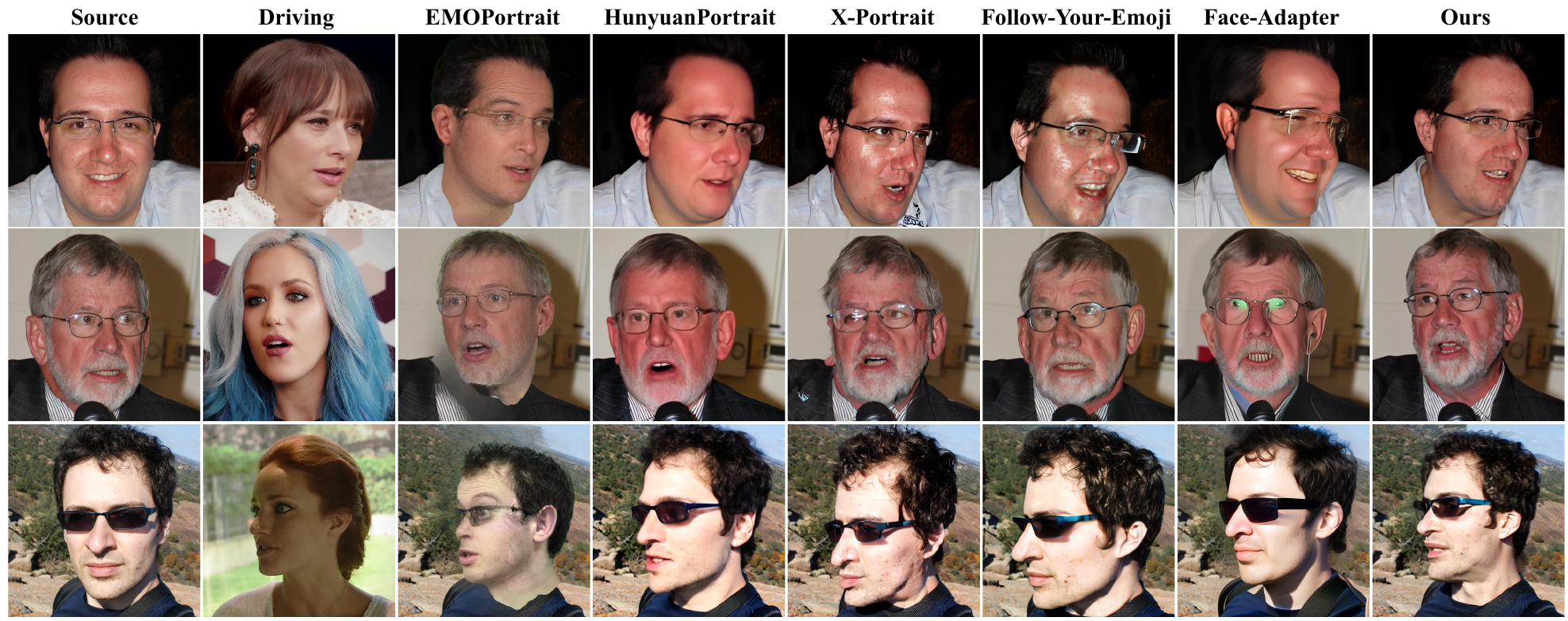}
\caption{Qualitative results on FFHQ dataset. ``Source'' and ``Driving'' denote the source and driving image, respectively.}
\label{ffhq_fig}
\vspace{-2mm}
\end{figure*}

Tab.~\ref{ffhq} and Fig.~\ref{ffhq_fig} show the quantitative and qualitative results on FFHQ dataset, respectively. 
We again achieve the best motion transfer while maintain relatively good source ID, which verifies good generalization ability of our model.

\begin{table}[t]
\centering
\caption{Ablation study of depth-guided sampling on the self-reenactment task.}
\label{ablate-depth}
\resizebox{\linewidth}{!}{
\begin{tabular}{p{1.5cm}<\centering|p{1.3cm}<\centering p{1.3cm}<\centering p{1.3cm}<\centering p{1.3cm}<\centering}
\toprule
Method      & LPIPS $\downarrow$ & CSIM $\uparrow$    & APD $\downarrow$ & AED $\downarrow$ \\ \midrule
w/o Depth & 0.213 & 0.770 & 9.659 & 0.730 \\
w/ Depth & \textbf{0.158} & \textbf{0.807} & \textbf{2.682} & \textbf{0.327} \\
\bottomrule
\end{tabular}
}
\vspace{-3mm}
\end{table}

\subsection{Ablation Study}
\noindent \textbf{Effectiveness of 3D Flow Encoding.}
Tab.~\ref{ablate-motion} shows the comparison of using different motion conditions as the input to the ControlNet. The first row is using driving landmark as the motion condition. It shows bad performance as the landmark lacks connection between the source and the driving motion.
\yt{
We also test the model using the predicted flow map \cite{shin2024instantdrag} as the motion condition to the ControlNet in the second row of Tab.~\ref{ablate-motion}.
It also performs poorly due to the difficulty of the flow estimation for each pixel under diverse motion and appearance changes.
In contrast, we directly compute the flows between the source and the target 3D head models, yielding geometrically consistent correspondences. This learning-free design ensures robust and accurate motion guidance across diverse subjects and poses.
}

\vspace{-2mm}
\begin{figure}[htp]
\centering
\includegraphics[scale=0.38]{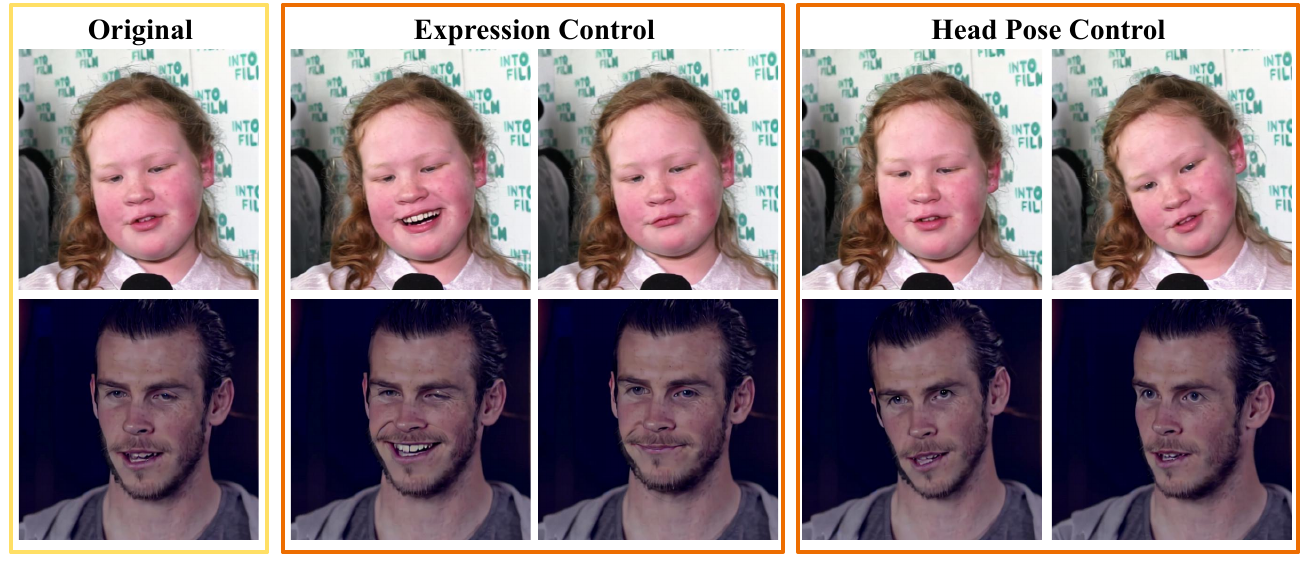}
\caption{Qualitative results on of expression and pose editing. "Original" denotes the original animation using the driving image. The second and third columns show expression control while the rest columns show head movement.}
\label{edit}
\vspace{-3mm}
\end{figure}

\noindent \textbf{Effectiveness of Depth-Guided Sampling.}
Tab.~\ref{ablate-depth} shows the ablation study of depth-guided sampling. ``w/o Depth ''denotes uniformly sampling $N$ points from $[d_{near}, d_{far}]$ as $P_{tgt}$. Compared to our model (``w/ Depth''), it performs poorly on all evaluation metrics, especially on the metrics that heavily rely on the correct motion guidance (AED and APD). It verifies that depth-guided sampling can capture more correct 3D flows corresponding to the 2D motion.

\subsection{User-specified Editing}

Fig.~\ref{edit} shows the qualitative result of editing during inference stage.
``Original" denotes the original animation using $I_{dri}$. The second and third columns show the animation of modifying the expression. The last two columns show the head pose control.
It verifies that our model can achieve diverse user-specified editing.

\section{Conclusion}
In this paper, we propose FG-Portrait, which takes the 3D flow as the new motion guidance for the diffusion-based portrait animation. The 3D flow establishes the motion correspondence between the source and the target portrait in the 3D space.
To utilize the 3D motion prior for the 2D animation task, we introduce 3D flow encoding as the new conditional input to the ControlNet. The flow encoding describes the corresponding 3D displacement for each target pixel moving back to the source location. 
We further introduce depth-guided sampling to improvement the alignment of the 3D flow encoding with the 2D motion change.
In addition, we support user-specified expression and head pose editing on the animated result during inference.
Extensive experiments verify the effectiveness of our model.

\vspace{1mm}
\noindent \textbf{Limitation.}
It has been observed that the 3D head models may struggle to represent the fine-grained expression due to the limited mesh resolution \cite{cudeiro2019capture, fan2022faceformer, feng2021learning, retsinas20243d}. 
In the future, we plan to investigate more advanced 3D head models for better portrait animation.

\clearpage
\maketitlesupplementary

\appendix
\section{Additional Ablation study}
\begin{figure}[h]
\centering
\includegraphics[scale=0.45]{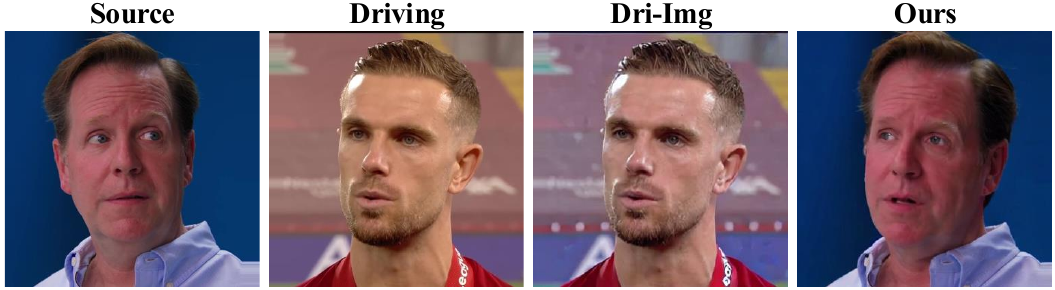}
\caption{Qualitative comparison between `Dri-Img' and Ours. 
`Dri-Img' directly copies the driving image as the output. In contrast, we can correctly transfer the driving motion to the source person.
}
\label{motion-ablation-figure}
\end{figure}

\noindent \textbf{Additional Ablation on Motion Condition.}
Tab.~\ref{supp-ablate-motion} shows comparison with using driving image as motion condition.
Although it shows much better results on the APD and AED, it cheats in the animation by simply copying the driving image as the final output as shown in the Fig.~\ref{motion-ablation-figure}.
In contrast, we can correctly transfer the motion and maintain the source identity.

\begin{table}[h]
\centering
\caption{Comparison of different motion conditions. `Dri-Img' denotes using driving image as the motion condition.  `S-APD' and `S-AED' are the APD and AED for self-reenactment task. `C-APD' and `C-AED' are the APD and AED for cross-reenactment task.}
\label{supp-ablate-motion}
\resizebox{\linewidth}{!}{
\begin{tabular}{p{1.5cm}<\centering|p{1.5cm}<\centering p{1.5cm}<\centering p{1.5cm}<\centering p{1.5cm}<\centering}
\toprule
Model     & S-APD $\downarrow$ & S-AED $\downarrow$    & C-APD $\downarrow$ & C-AED $\downarrow$ \\ \midrule
Dri-Img & 1.060 & 0.131 & 1.216	& 0.144 \\
Ours & 2.682 & 0.327 & 7.764 &	0.652 \\
\bottomrule
\end{tabular}
}
\end{table}

\noindent \textbf{Ablation Study on $N$ and $\delta$ in the 3D Flow Encoding.}
Tab.~\ref{N} and Tab.~\ref{delta} show the ablation study of $N$ and $\delta$ on the self-reenactment of VFHQ.
The performance is generally robust to different combination of $N$ and $\delta$, with a slight degradation when using fewer samples ($N=10$) or a wider sampling range ($\delta=0.05$m), due to insufficient sampling density or less accurate 3D flow encoding.
Memory of generating 10-frame video mildly increases with larger $N$, while remain constant under different $\delta$. 

\begin{table}[h]
\centering
\caption{Ablation study of $N$ in the self-reenactment on VFHQ.}
\label{N}
\resizebox{\linewidth}{!}{
\begin{tabular}{c|ccccc}
\hline
Method                              & LPIPS $\downarrow$ & CSIM $\uparrow$ & APD $\downarrow$ & AED $\downarrow$ & Mem(MB)  \\ \hline
N=10    &  0.164   &  0.798  & 2.724     & 0.332  & 34110    \\ 
N=30   & 0.160  & \textbf{0.807}  & \textbf{2.540}  & 0.334 & 34708   \\
ours   & \textbf{0.158}  & \textbf{0.807}  & 2.682  & \textbf{0.327} & 34402    \\
\hline
\end{tabular}
}
\vspace{-3mm}
\end{table}

\begin{table}[h]
\centering
\caption{Ablation study of $\delta$ in the self-reenactment on VFHQ.}
\label{delta}
\resizebox{\linewidth}{!}{
\begin{tabular}{c|ccccc}
\hline
Method                              & LPIPS $\downarrow$ & CSIM $\uparrow$ & APD $\downarrow$ & AED $\downarrow$ & Mem(MB) \\ 
\hline
 $\delta=0.05m$    & 0.162 & 0.803  & 2.742  & 0.330  & 34402     \\ 
$\delta=0.005m$   & 0.160 & 0.804 & \textbf{2.641} & \textbf{0.326}  & 34402   \\
ours   & \textbf{0.158}  & \textbf{0.807}  & 2.682  & 0.327  & 34402   \\
\hline
\end{tabular}
}
\end{table}

\section{More Qualitative Comparison with Baselines}
\begin{figure*}[h]
\centering
\includegraphics[scale=0.5]{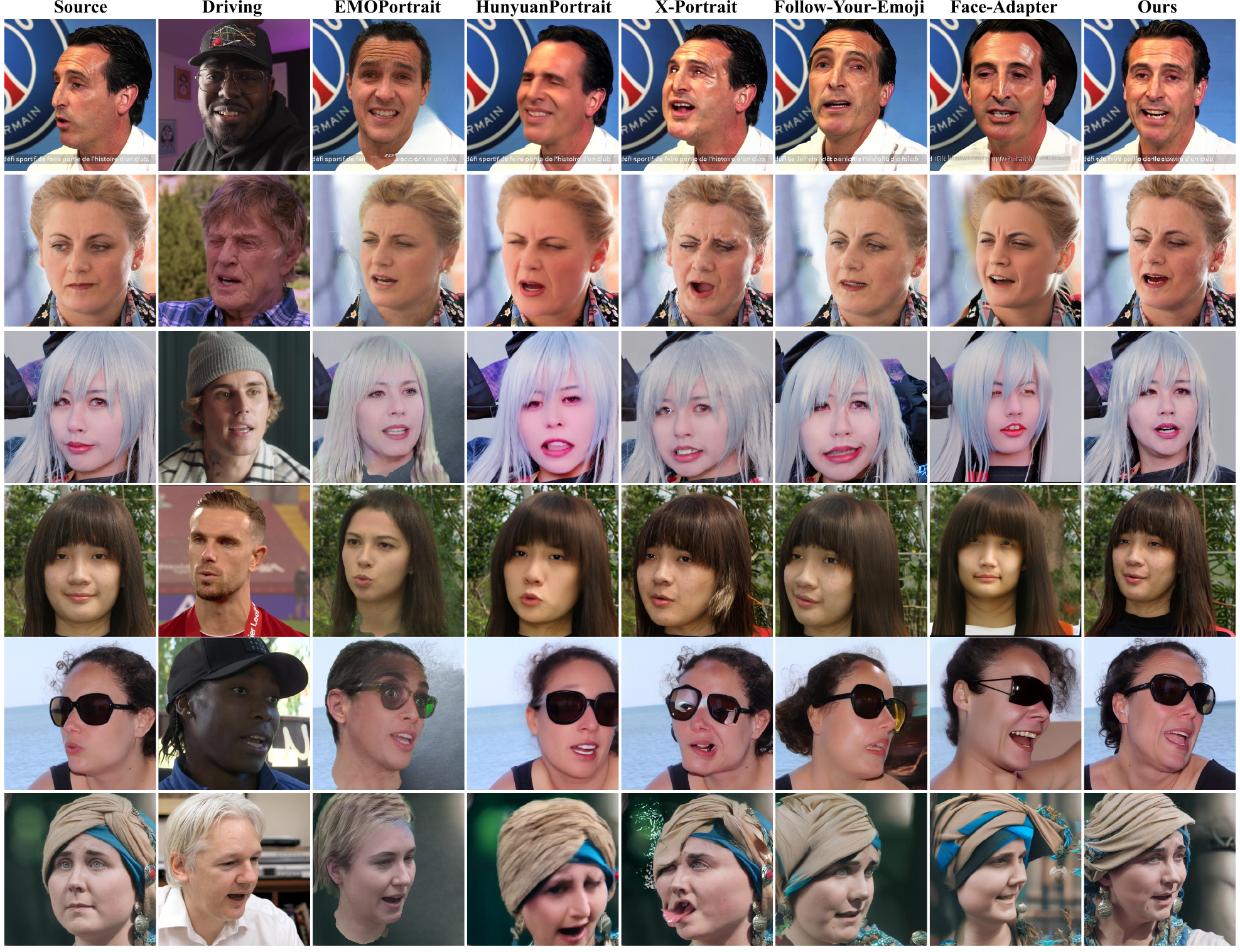}
\caption{Qualitative comparisons on cases with diverse motion and appearance change.}
\label{supp_vfhq}
\end{figure*}

Fig.~\ref{supp_vfhq} shows more qualitative comparisons on testing samples with diverse motion and appearance variations. We show better motion transfer and maintain source identity under these challenging scenarios, \eg identities with long hair, complex accessories, different ethnicities and ages. 

\begin{figure*}[h]
\centering
\includegraphics[scale=0.8]{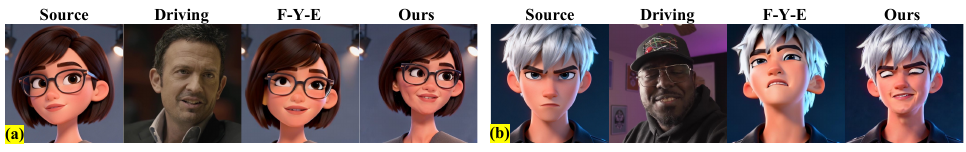}
\caption{Qualitative comparisons on cartoon portraits. (b) shows a challenging case, which is discussed in Sec.~\ref{sec:cartoon}.}
\label{cartoon}
\end{figure*}

\section{Animation with Cartoon Portrait}
\label{sec:cartoon}
Fig.~\ref{cartoon} presents the results on cartoon portraits, where ``F-Y-E'' denotes Follow-Your-Emoji model. Compared to the baseline, we can more accurately drive the cartoon head.
We notice that there is artifact of eyelid closure in Fig.~\ref{cartoon} (b).
The reason is that our model lacks appropriate appearance priors for cartoon portraits since it is trained exclusively on realistic human portraits. It can be addressed by finetuning on cartoon-specific datasets.

\section{Video Results}
We provide video results in the supplementary material. Compared to SOTA methods, we can maintain good temporal consistency and show superior motion transfer at the same time.

\section{Temporal consistency Analysis}
Tab.~\ref{fvd} shows the temporal consistency analysis on the self-reenactment task of VFHQ. We use Frechet Video Distance (FVD), which is the lower the better temporal consistency.
We achieve the second lowest FVD score, which verifies good temporal consistency.

\begin{table}[h]
\centering
\caption{Temporal consistency analysis on the self-reenactment task of VFHQ. Best result is marked bold. `FYE' and `Hunyuan' are short for Follow-Your-Emoji and HunyuanPortrait, respectively.}
\label{fvd}
\resizebox{\linewidth}{!}{
\begin{tabular}{c|cccccc}
\hline
Method & \small{EMOPortrait} & \small{X-Portrait} & \small{FYE} & \small{Face-Adapter} & \small
{Hunyuan} & Ours \\ \hline
  FVD     &  567.2           &  575.3          &  \textbf{382.6}                 &   472.3           &  430.2               & 412.1     \\ \hline
\end{tabular}
}
\end{table}

\newpage
{
    \small
    \bibliographystyle{ieeenat_fullname}
    \bibliography{main}
}


\end{document}